\documentclass[journal]{IEEEtran}

\usepackage{times}
\usepackage{epsfig}
\usepackage{graphicx}
\usepackage{amsmath}
\usepackage{amssymb}
\usepackage{subfiles}
\usepackage{multirow} 
\newcommand{\ra}[1]{\renewcommand{\arraystretch}{#1}}
\usepackage{booktabs}
\usepackage[ruled]{algorithm2e}

\makeatother

\usepackage{makecell}
\usepackage[pagebackref=true,breaklinks=true,colorlinks,bookmarks=false]{hyperref}
\ifCLASSINFOpdf
\else
\fi
\begin{document}
\title{Robust Face Anti-Spoofing with \\ Dual Probabilistic Modeling}

\author{Yuanhan~Zhang,
        Yichao~Wu,
        Zhenfei~Yin,
        Jing~Shao,
        and~Ziwei~Liu,~\IEEEmembership{Member,~IEEE}% <-this % stops a space
% \thanks{Yuanhan~Zhang and Yichao~Wu have equal contribution.}
\thanks{Yichao~Wu, Zhenfei~Yin and Jing~Shao are with SenseTime Research.}
\thanks{Yuanhan~Zhang and Ziwei~Liu are with Nanyang Technological University.}}% <-this % stops a space
\maketitle

% \begin{abstract}
% The abstract goes here.
% \end{abstract}

% % Note that keywords are not normally used for peerreview papers.
% \begin{IEEEkeywords}
% IEEE, IEEEtran, journal, \LaTeX, paper, template.
% \end{IEEEkeywords}

\begin{abstract}
The field of face anti-spoofing (FAS) has witnessed great progress with the surge of deep learning. 
Due to its data-driven nature, existing FAS methods are sensitive to the noise in the dataset, which will hurdle the learning process.  
However, very few works take the noise modeling into consideration in FAS.
In this work, we attempt to fill this gap by automatically addressing the noise problem from both label and data perspectives in a probabilistic manner. 
Specifically, we propose a unified framework called \textbf{Dual Probabilistic Modeling} (DPM), with two dedicated modules, \textbf{DPM-LQ} (Label Quality aware learning) and \textbf{DPM-DQ} (Data Quality aware learning).
Both modules are designed based on the assumption that data and label should form coherent probabilistic distributions. 
DPM-LQ is able to produce robust feature representations without overfitting to the distribution of noisy semantic labels.
DPM-DQ can eliminate data noise from `False Reject' and `False Accept' during inference by correcting the prediction confidence of noisy data based on its quality distribution.
Both modules can be incorporated into existing deep networks seamlessly and efficiently. 
Furthermore, we propose the generalized DPM to address the noise problem in practical usage without the need of semantic annotations.
Extensive experiments demonstrate that this probabilistic modeling can 1) significantly improve the accuracy, and 2) make the model robust to the noise in real-world datasets.
Without bells and whistles, our proposed DPM achieves state-of-the-art performance on multiple standard FAS benchmarks.

\end{abstract}


\section{Introduction}

Face interaction systems have become an essential part in real-life applications, with the successful deployments in electronic identity authentication. 
Meanwhile, it is challenging to deal with Presentation Attacks (PA)~\cite{PA1} in practical usage. 
In order to protect our privacy and property from being illegally used by others, 
Face Anti-Spoofing (FAS)~\cite{PA2,FAS1,AtoumFaceAU}, which aims to determine whether a presented face is an attacker or client, has emerged as a crucial technique and attracted extensive interests in recent years~\cite{Galbally2014FASsurvey}.

% In the past decades, FAS algorithms based on deep learning have been great progress. 
% On the one hand, approaches based on texture analysis have achieved great success~\cite{texture1,texture2,yztHMP,yztCDC}. 
% Those methods focus on introducing special texture differences including color distortions, Moir patterns and \textit{etc.} from input data. On the other hand, researches focus on leverage side information~\cite{lyjauxuliary,AtoumFaceAU,yztHMP,CelebA-Spoof,KimBASN}, 
% such as geometric information and semantic information to improve binary classification. 

%In the past decades, FAS algorithms have achieved great progress.
%On the one hand, 
Traditional FAS algorithms mainly focus on devising hand-crafted descriptors to capture discriminated features~\cite{HSV:boulkenafet2016face,Patel2016SIFT}.
Nowadays, great breakthroughs have been witnessed in the field of FAS based on deep learning methods, either by the better architecture designs~\cite{yztCDC,yztHMP} or utilizing the side information~\cite{KimBASN,CelebA-Spoof,liu20163d}.
%(\textit{e.g.}, color distortions, Moir patterns and \textit{etc.},
%which imply detailed information to classify live and spoof faces.
%zyh
% Both traditional methods and deep learning based methods are committed to fully leveraging information in FAS datasets.
%
% As a data-driven approach, 
Given deep learning methods are sensitive to the noise in datasets~\cite{zhang2016understanding}, 
% and the FAS datasets become larger, 
there is an increasing demand for developing robust learning frameworks. 
However, as far as we know, very few approaches take the noise of datasets into consideration in the field of FAS.

\begin{figure}[t]
\centering
\includegraphics[width=0.45\textwidth]{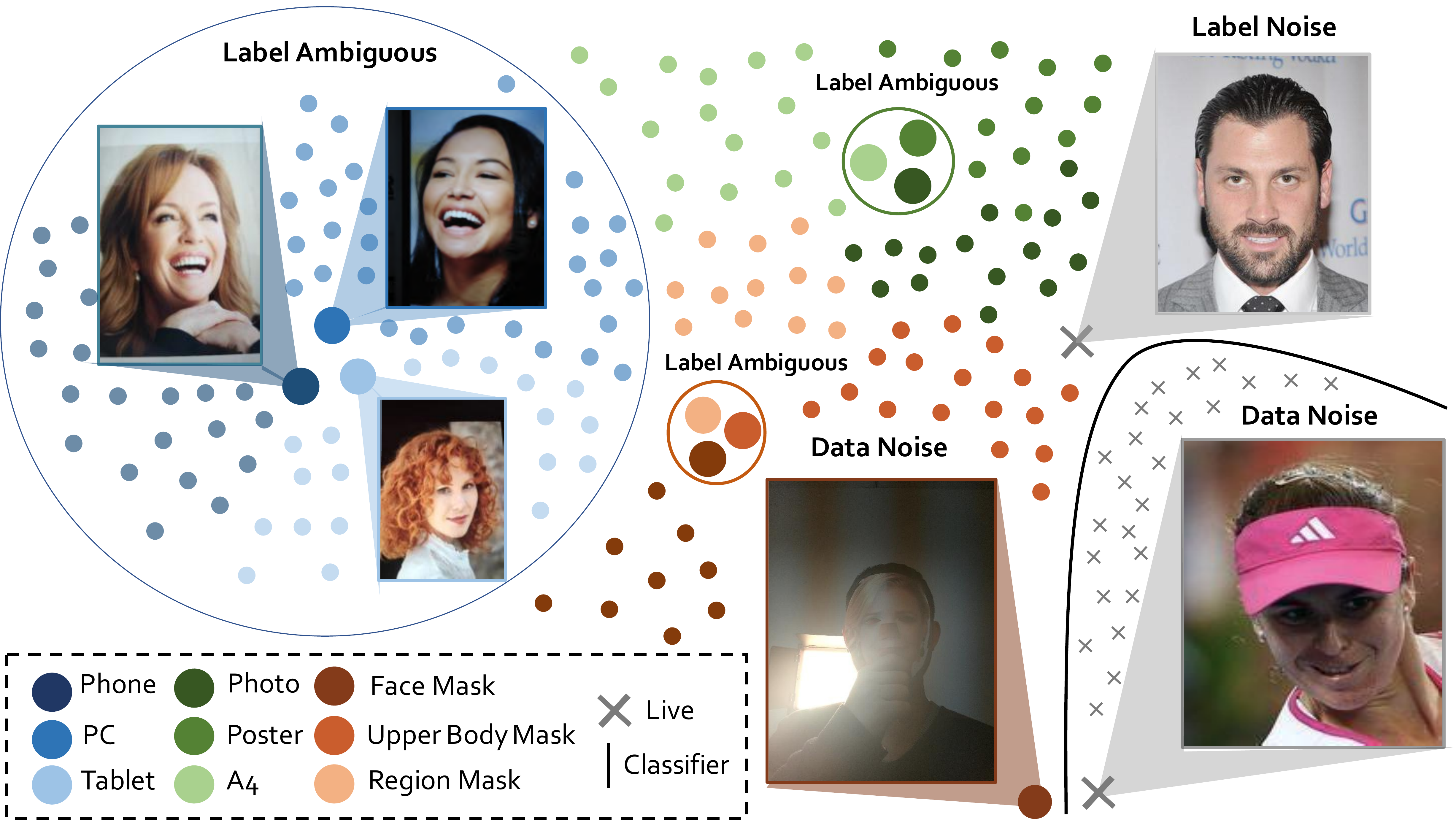}
\caption{\textbf{Three kinds of noise widely exist in face anti-spoofing datasets}: \textit{label ambiguous}, \textit{label noise} and \textit{data noise}. Specifically. \textit{Label ambiguous} refers to the fact that it is hard to assign a specific semantic labels to some input data.
We term \textit{label noise} as wrongly annotated data. 
\textit{Data noise} refers to images of extremely low quality, such as the severely obscured face.
}.
%
% \vspace{-13pt}
\label{figure:fig1}
\end{figure}

As shown in Fig.~\ref{figure:fig1},~\footnote{In our experiments, all the images presented in figures shown in this paper are cropped by its face area.} we conduct in-depth investigations on the noise in typical FAS datasets.
%A typical visualization of FAS datasets is shown in Figure~\ref{figure:fig1}. %
From our observation, there exist mainly three types of noise in common datasets: \textit{label ambiguous}, \textit{label noise}, and \textit{data noise}. 
% Specifically, this noise roughly takes up 8\%, 5\%, 5\%, respectively in CelebA-Spoof~\cite{CelebA-Spoof}.
\textbf{1)} \textit{Label ambiguous} refers to the fact that it is hard to assign a specific semantic label to input data, which widely exist in FAS datasets containing rich annotations, limiting the effectiveness of the side information.
Although the auxiliary semantic information has demonstrated successfully in previous work~\cite{CelebA-Spoof},
the attribute boundary of different spoof types (such as \textit{Phone, Tablet, and PC} in Fig.~\ref{figure:fig1}) is not always clear. 
\textbf{2)} We term \textit{label noise} as wrongly annotated data, which is inevitable in large-scale benchmarks.
Generally, if models converge to the sub-optimal solution on the training set with inaccurate labels, there would be large performance gaps on the test set. 
\textbf{3)} \textit{Data noise} refers to images of extremely low quality, such as the severely obscured face. 
Usually, common models cannot generalize to deal with such cases; and we find noisy data takes up a large portion of False Reject and of False Accept during inference, hindering further performance improvement.
Here we aim to tackle the following problem: 
\textit{How to boost FAS by automatically dealing with data and label noise in a unified framework?}

Motivated by the above analysis, 
we propose a clean yet powerful framework 
called \textbf{Dual Probabilistic Modeling} (DPM), which can effectively address the impact of noise in FAS datasets
% %. Different from pioneer works, 
% in this paper, 
% %
% the proposed DPM attempts to boost the potential of datasets 
from the perspective of probabilistic modeling.
DPM consists of \textbf{DPM-LQ} (Label Quality aware learning) and \textbf{DPM-DQ} (Data Quality aware learning).
Both modules are based on the assumption of coherent probabilistic distribution, and can be learned in a unified framework.  
For DPM-LQ,
we assume that noise existing in the semantic feature representations conforms to the Gaussian distribution.
In order to prevent the model from overfitting to distribution of inaccurate labels, 
we disentangle the mid-level representations into two parts, 
which are the robust feature embedding and label uncertainty in the latent space.
Thus,
through adequate training, 
the classification model can generate the robust semantic feature representation for each sample. 
On the other hand, 
for DPM-DQ, 
we suppose the distribution of the feature space of the whole sample as another Gaussian distribution, 
where Live/Spoof class center is the mean of the distribution, 
the degree of data quality can be modeled as the variance. 
In this way, 
by correcting the prediction confidence of noisy data based on its quality, 
noisy data can be eliminated from False Reject and from False Accept during inference.

Indeed, re-weighting~\cite{sukhbaatar2014training} and curriculum learning~\cite{curriculum_learning} are widely used to ease the impact of label noise. However, they can barely solve the problem of label ambiguous which is unique in FAS datasets.
Besides, quality assessment models (QAM) ~\cite{best2018learning} can filter the noisy data. 
Unfortunately, it is necessary to acquire ``clean data'' to get a good QAM. 
% Furthermore, the system may become unstable and time-consuming because of the additional modules.
%
Our proposed DPM has three appealing properties compared to the prior noise-resilient methods. 
\textbf{1)} Firstly, the problem of the label noise and label ambiguous can be solved simultaneously under a single probabilistic model, which is especially effective for the FAS task.
\textbf{2)} Secondly, the data quality modeling of each image can be incorporated implicitly during training without extra models, 
which can certainly make DPM both stable and efficient.  
\textbf{3)} Thirdly, DPM can further improve the model performance on the test set which is already cleaned by QAM, which indicates the functions of QAM and DPM are not completely overlapped.

% Generally, 
% the proposed DPM can improve the FAS performance 
% by increasing the potential of data utilization. 
% %
% Furthermore, 
% as most academic (such as Oulu-NPU~\cite{oulu-npu}, SiW~\cite{lyjauxuliary}) and industrial datasets lack sufficient semantic information,
% %
% we introduce a generalized version of DPM which can annotate semantic labels by our self-distributed algorithms.

In summary, the contributions of this paper are three-fold:
\begin{itemize}
\item [1)] 
We comprehensively study the noise problem in FAS datasets for the first time. A unified framework called Dual Probabilistic Modeling (DPM) is proposed. DPM consists of DPM-LQ and DPM-DQ to ease the negative impact of noise from both label and data perspectives.
\item [2)]
We further design generalized DPM to tackle real-world FAS datasets without the need of semantic annotations. It successfully deals with the noisy labels and degraded data within large-scale samples.
\item [3)]
Extensive experiments demonstrate that DPM, without bells and whistles, can achieve state-of-the-art results on multiple standard FAS benchmarks.
\end{itemize}

\section{Related Works}

% \begin{figure}[t]
% \centering
% \includegraphics[width=0.45\textwidth]{Dual_Probabilistic_Modeling/figure/frr_explanation_compressed.pdf}
% \caption{Based on modeled data quality, the prediction confidence of noise data drop, even smaller than live data which original confidence is smaller than them, which can further improve FRR.}
% \label{figure:frr_explanation}
% \end{figure}

\begin{figure*}[t]
\centering
\includegraphics[width=\textwidth]{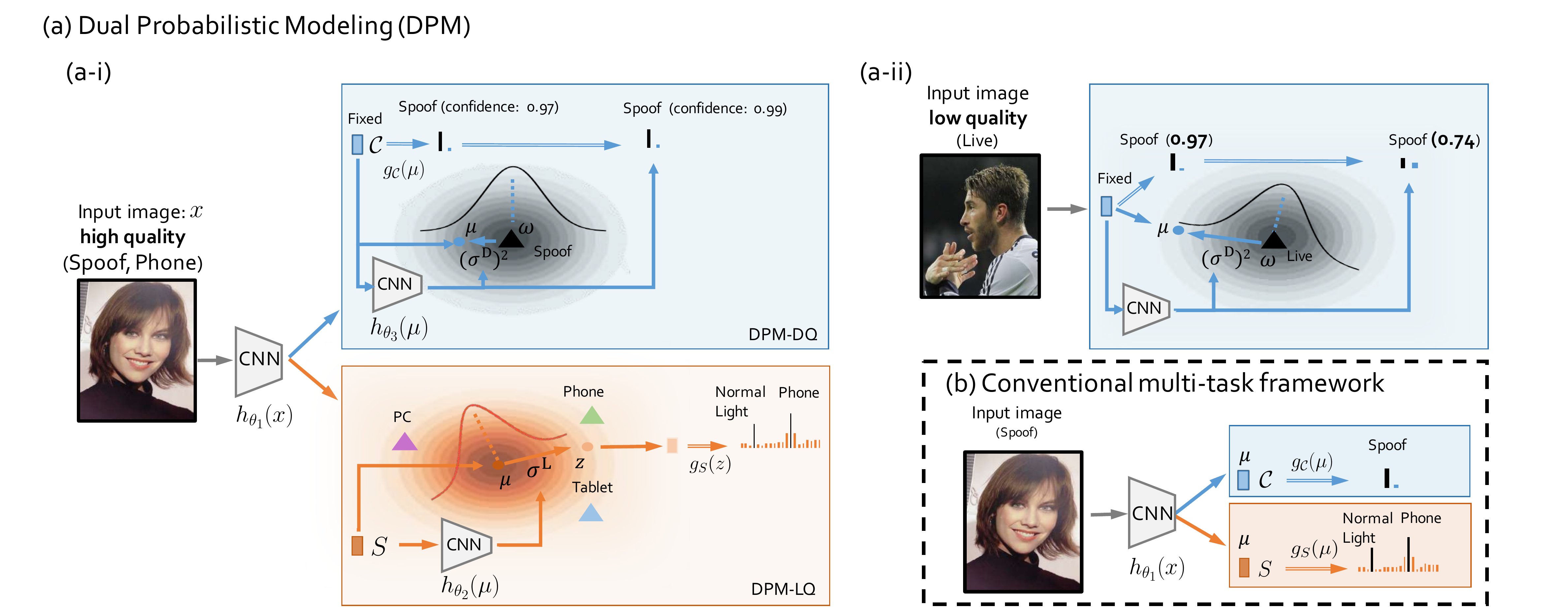}
\caption{\textbf{Overview of Dual Probabilistic Modeling (DPM)}. DPM-LQ can be conducted on any semantic labels. Triangles represent the class center. 
(a-i) For input data that labeled Phone class, it is hard to embed it to a specific spoof type label distribution: (PC, Tablet or Phone), which implies label ambiguous. Rather than training feature extractor $h_{\theta_{1}}(\cdot)$ to embed its feature representation $\mu$ to a specific representation, a Gaussian distribution of this input data is formed in latent space by DPM-LQ, and the cross-entropy loss is calculated based on the $z$ which is sampled from this distribution.
%
% As a result, a Gaussian distribution of this input data is formed in latent space, $\mu$ is mean of this distribution and $\sigma^\textit{L}$ is the standard deviation, $z$ is sampled from this Gaussian distribution for calculating classification loss.
%
% Since the classification loss is calculated based on the $z$ instead of $\mu$, formed Gaussian distribution prevents the feature extractor $h_{\theta_{1}}(\cdot)$ from overfitting to data with label ambiguous, hence achieving better class separability and better generalization.
%
(a-ii) DPM-DQ can predict data quality by formulating another two Gaussian distributions among Live/Spoof class center in the latent space. The $(\sigma^\textit{D})^2$ of low quality data is only related to the Gaussian distribution of spoof label. Since low quality data has a larger $(\sigma^\textit{D})^2$, $(\sigma^\textit{D})^2$ can be used to correct the prediction confidence of low quality data. (b) A conventional FAS model based on multi-task fashion.
}
% \vspace{-13pt}
\label{figure:CNN}
\end{figure*}

\noindent \textbf{Face Anti-Spoofing.}
% Most traditional methods focus on devising hand-crafted features from the faces, such as HSV~\cite{HSV:boulkenafet2016face}, LBP~\cite{texture1,Pereira2012LBP,Pereira2013LBP,Mtt2011LBP}, SIFT~\cite{Patel2016SIFT}, HOG~\cite{Komulainen2013HoG,Yang2013HoG}, SURF~\cite{Boulkenafet2017SURF} and Fourier spectrum~\cite{Fourier:li2004live}. Based on these features, SVM or LDA~\cite{Pereira2012LBP} can be used as a classifier to discern live or spoof. 
With the development of deep learning, researchers resort to Convolutional Neural Network (CNN) for face anti-spoofing. Compared to training CNN to learn a binary classifier, auxiliary supervision is widely used to further improve the performance of binary classification supervision. Atoum \textit{et al.}~\cite{AtoumFaceAU} leverage the Fully Convolutional Network (FCN) to assistant the binary classification through depth map. Liu \textit{et al.}~\cite{Liu_2018_ECCV,liu20163d} propose remote toplethysmography (rPPG signal)-based methods to foster the development of 3D face anti-spoofing. Liu \textit{et al.}~\cite{lyjauxuliary} combine the rPPG signal and depth map to further improve the performance of FAS. Kim \textit{et al.}~\cite{KimBASN} leverage depth map and reflection map as the bipartite auxiliary supervision. In addition to geometric information including depth map and reflection map, Zhang \textit{et al.}~\cite{CelebA-Spoof} leverage semantic information to help Live/Spoof classification which achieves better performance. Further, to extract rich intrinsic features among live faces and various kinds of spoof types, Yu \textit{et al.} devise bilateral convolutional network~\cite{yztHMP} and central difference convolution~\cite{yztCDC} for capturing intrinsic detailed patterns of faces. Even though FAS approaches mentioned above are committed to fully leverage information including data and its label in the FAS datasets, few works take noise in FAS datasets into consideration.

\noindent \textbf{Label and Data Noise in Deep Learning.}
% With the advancement of deep neural network, image classification performance makes a big leap. Importantly, current image classification methods are fueled by the availability of large-scale dataset. However, 
Label noise and data noise are common problems in large-scale datasets~\cite{algan2019labelnoisesurvey,kendall2017uncertainties}. Several methods are proposed to ease the negative impact of label noise and data noise in datasets. 
Sukhbaatar \textit{et al.}~\cite{sukhbaatar2014training} explicitly model label noise through assigning weight as the importance of each sample. 
Curriculum learning~\cite{curriculum_learning} proposes to start from clean labeled data and go through noisily
labeled data to guide training.  Besides, dropout~\cite{dropout} are proposed to boost the model performance by regularizing parameters.
% can also be used to avoid the impact of label noise. 
On the other hand, Kendall \textit{et al.}~\cite{kendall2017uncertainties} propose data noise as one kind of aleatoric uncertainty. 

As far as we know, we are the first to propose a network which can ease both label noise and data noise in a unified structure in the field of face anti-spoofing. Compared to distribution modeling methods~\cite{datauncertainty,PFE} for easing noisy data in other fields, the proposed DPM has two appealing advantages specifically designed for FAS. 1) DPM models the distributions for both main task and auxiliary task of FAS. 2) The generalized DPM proceeds distribution modeling under the self-distributed setting, as it is both expansive and laborious to acquire semantic labels.

\section{Dual Probabilistic Modeling}
\label{section:methodology}
% Inspired by AENet~\cite{CelebA-Spoof}, Dual Probabilistic Modeling(DPM) built upon multi-task fashion. 
%

Noisy data and inaccurate labels are crucial problems in FAS. To deal with these problems in a unified framework, we propose Dual Probabilistic Modeling (DPM) including DPM-LQ and DPM-DQ. As shown in Fig.~\ref{figure:CNN} (a), two different Gaussian distributions are formed by DPM, which aims to ease the problems of both noisy data and inaccurate labels.
% both modules are based on the assumption of coherent probabilistic distribution, and can be learned in a unified framework.
%
In this section, we firstly present two probabilistic modelings of DPM in detail,
then we generalize DPM to be applicable in most academic datasets or real-world scenarios which lack sufficient semantic information.
%
% Besides, notations used are shown in \textit{Appendix}. %Table~\ref{table:notations}.
\subsection{Notations}
\label{sec:framework}
Here, we list all notations we used in this paper.
\setlength{\tabcolsep}{5pt}
\begin{table}[h!]
\footnotesize
\centering
\ra{1.2}
% \caption{Notations}
\label{table:notations}
% \resizebox{0.5\textwidth}{!}{%
\begin{tabular}{@{}ll@{}}
\Xhline{1pt}
Notation & Meaning \\ \hline
$c_i$ & The Live/Spoof Label of $x_i$ \\
$\mathcal{D}^{def}$ & The semantic label deficient dataset \\
$\mathcal{D}^{suf}$ & The semantic label sufficient dataset \\
$h_{\theta_{1}}(\cdot)$ & The feature representation extractor \\
$h_{\theta_{2}}(\cdot)$ & The variance embedding network \\
$g_{\mathcal{C}}(\cdot)$ & The Live/Spoof label classifier \\
$g_{\mathcal{S}}(\cdot)$ & The semantic label classifier \\
$\mu_i$ & The feature representation of $x_i$ \\
$\omega_\mathcal{S}$ & The parameter matrix of $g_{\mathcal{S}}(\cdot)$ \\
$\omega_{\mathcal{S}_i}$ & One row vector of $\omega_\mathcal{S}$, which depends on the $s_i$ \\
$\omega_\mathcal{C}$ & The parameter matrix of $g_{\mathcal{C}}(\cdot)$ \\
$\omega_{\mathcal{C}_i}$ & One row vector of $\omega_\mathcal{C}$, which depends on the $c_i$ \\
$(\sigma^\text{L}_i)^2$ &  The variance of the Gaussian distribution of $\mu_i$ in DPM-LQ \\ 
$(\sigma^\text{D}_i)^2$ &  The variance of the Gaussian distribution of $\omega_{\mathcal{C}_i}$ in DPM-DQ \\ 
$\mathcal{S}$ & A set of semantic labels from one category \\
$s_i$ &  The semantic label of category $\mathcal{S}$ in $x_i$  \\
% $s_i$ & The semantic label of one category of i-th input data \\
$\mathcal{S}^\text{a}$ & A set of annotated semantic labels \\
$\mathcal{S}^\text{s}$ &  A set of self-distributed semantic labels \\
$x_i$ & Given the i-th input data \\
$z_i$ & The sampled semantic feature representation of $x_i$ \\
% $ x_i $ & The Live/Spoof label of i-th input data \\

% $\mathcal{G}$ & Geometric Label \\
\Xhline{1pt}
\end{tabular}%
% }
\end{table}

\subsection{DPM-LQ}
\label{sec:framework}
% \noindent \textbf{Inaccurate Label Aware Learning (DPM-LQ).}
As shown in Fig.~\ref{figure:CNN} (b), 
for the conventional multi-task model based on semantic information in FAS, 
there are normally three modules: feature extractor $h_{\theta_{1}}(\cdot)$, Live/Spoof classifier $g_{\mathcal{C}}(\cdot)$, and semantic information
% \footnote{For simplicity of formula, we only leverage semantic spoof type $\mathcal{S}^{\text{s}}$ to represent semantic information.} 
classifiers $g_{\mathcal{S}}(\cdot)$.
% , where $\mathcal{S}^{\text{s}}$ represents \textbf{s}emantic \textbf{s}poof type.
%
With the i-th input image $x_i \in \mathcal{X}$, its Live/Spoof label $c_i \in \mathcal{C}, \mathcal{C}= \left \{ 0,1 \right \}$ , and its semantic label.\footnote{\scriptsize{That is, spoof types and illumination conditions. For the simplicity of notation, we use $\mathcal{S}$ to indicate the set of semantic labels from one category.}} $s_i \in \mathcal{S}, \mathcal{S}= \left \{ 0,1,...,A \right \}$,
the model is trained by minimizing cross-entropy loss between $g_{\mathcal{C}}(h_{\theta_{1}}(x_i))$ and ${c_i}$, $g_{\mathcal{S}}(h_{\theta_{1}}(x_i))$ and ${s_i}$, $\mu_i$ predicted by $h_{\theta_{1}}(x_i)$ is the feature representation of $x_i$.

% Feature representation $h_{\theta_{1}}(x_i)$ of 
Conventionally, $\mu_i$ will be fed into $g_{\mathcal{S}}(\cdot)$ for the cross-entropy loss of the semantic label: $\mathcal{L}_{\mathcal{S}}$ as follows:
\begin{equation}
\label{equation:conventional_loss}
    \mathcal{L}_{\mathcal{S}} = \frac{1}{N}\sum_{i}^{N}-\log \frac{e^{\omega_{s_i}\mu_{i}}}{\sum_{\mathcal{S}}^{}e^{\omega_{\mathcal{S}}\mu_{i}}},
\end{equation}
where $N$ is the number of training data, $\omega_\mathcal{S} \in \mathbb{R} ^{A\times B}$ is the parameters of $g_{\mathcal{S}}(\cdot)$, $A$ refers to the numbers of classes of semantic information and $B$ refers to the dimension of each row vector of $\omega_\mathcal{S}$, $\omega_{s_i}$ is one row vector of $\omega_\mathcal{S}$, which depends on the semantic label $s_i$. Optimization of the Eq.~\ref{equation:conventional_loss} inclines to closer the distribution of $\omega_{s_i}$ to the distribution of $\mu_{i}$. However, if $\mu_{i}$ is annotated by inaccurate labels, overfitting to its representation distribution might cause performance gaps between training set and test set.

\noindent \textbf{Distributional Modeling.} 
To ease the impact of inaccurate labels, DPM-LQ (see Fig.~\ref{figure:CNN} (a-i)) implicitly models a distribution over the semantic feature representation $\mu_i$ as follows:
% In the training phase, we assume that the semantic feature representation of $\mathcal{X}_{i}$ is sample from a Gaussian distribution.  
%
\begin{equation}
    p(z_{i} \mid x_{i}) = \mathcal{N}(z_i;\mu_{i},({\sigma_{i}^{\text{L}}})^{2}\text{I}),
\end{equation}
where $\mu_{i} = h_{\theta_{1}}(x_{i})$, $\sigma_{i}^{\text{L}} = h_{\theta_{2}}(\mu_{i})$, 
$\theta_{1}$ and 
$\theta_{2}$ represent model parameters \textit{w.r.t} $\mu_{i}$ and $\sigma_{i}^{\text{L}}$. 
Specifically, for a training sample $x_i$, DPM-LQ leverages the \textit{probabilistic} semantic feature representation $z_i$ which is sampled from the Gaussian distribution rather than the \textit{deterministic} semantic feature representation $\mu_i$ for the calculation of $\mathcal{L}_{\mathcal{S}}$.

\noindent \textbf{Semantic Information Classification Loss.} 
Different from the Eq.~\ref{equation:conventional_loss}, cross-entropy loss of DPM-LQ calculates the distribution difference between  $z_i$ and $\omega_{s_i}$ rather than $\mu_i$ and $\omega_{s_i}$ as follows:

\begin{equation}
\label{equation:label_uncertainty_loss}
    \mathcal{L}_{\mathcal{S}} = \frac{1}{N}\sum_{i}^{N}-\log \frac{e^{\omega_{s_i}z_{i}}}{\sum_{\mathcal{S}}^{}e^{\omega_{\mathcal{S}}z_{i}}}.
\end{equation}
However, since the nature of the sampling operation is not differential, the back propagation of the gradient flow during model training would be prevented.
Therefore, for $z_{i}$ which is a random sample from a Gaussian distribution, we resort to the re-parameterization trick~\cite{Kingma2014AutoEncodingVB} to ``sample'' the $z_{i}$ from the distribution. 
Formally, for a $\mu_i$, we sample an independent random noise $\varepsilon$ from the normal distribution and formalize the probabilistic semantic feature representation as follows:
\begin{equation}
    z_{i} = \mu_{i} + \varepsilon *\sigma_{i}^{\text{L}}, \quad \varepsilon\sim \mathcal{N}(0,\text{I}).
\end{equation}
Thus, we turn the random part of $z_{i}$ to random noise $\varepsilon$ and make $\mu_{i}$ and $\sigma_{i}^{\text{L}}$ trainable.
%
% Last, because $\sigma_{i}^{\text{L}}$ prevents semantic feature representation $\mu_{i}$ from overfitting to coarse label distribution.  
% %
% $\mu_{i}$ can represent the robust semantic feature representation.
%
Specifically, if a given training sample has an inaccurate label, instead of forcing distribution of $\omega_{s_i}$ fitting to the distribution of $\mu_{i}$, DPM-LQ calculates a larger variance to ``give up'' this sample, which reduces its influence on the distribution of $\omega_{s_i}$. In other words, this extra dimension allows the model to focus more on data with clean labels rather than samples with inaccurate labels, thereby achieving better class separability and better generalization.

\begin{figure*}[t]
\centering
\includegraphics[width=0.92\textwidth]{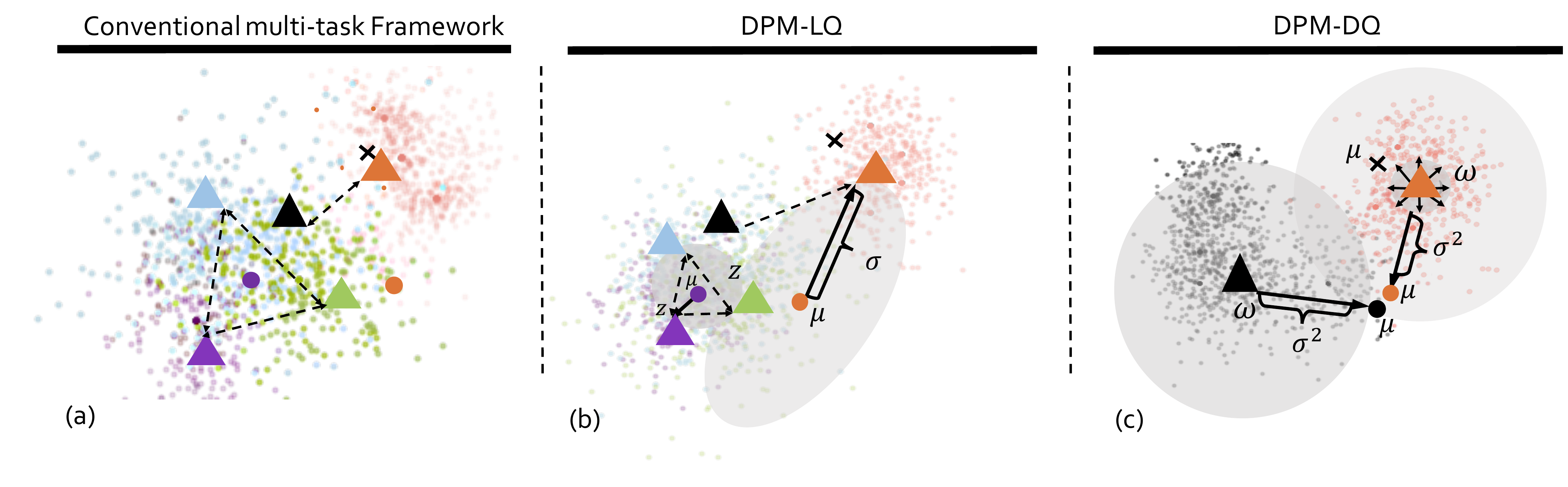}
% \vspace{-15pt}
\caption{\textbf{Feature distribution visualizations of test data on the CelebA-Spoof intra dataset test using t-SNE~\cite{tsne}}. \textbf{Color Indicate:} \textit{Blue}: PC, \textit{Green}: Tablet, \textit{Purple}: Phone, \textit{black}: Spoof,
\textit{Orange}: Live. 
\textbf{Shape Indicate}: \textit{Triangle}: Class Center. \textit{Cross}: Spoof Hard Case. 
\textit{Dot}: Sample Data.
% \textit{Dotted Line}: Distances.
% \textit{Solid Line}: Variances of Gaussian distributions. 
(a) The distribution of features representation bases on the the conventional multi-task model. It is hard to embed the purple dot (data with label ambiguous) to a specific spoof type label distribution, which causes model to converge slowly. The orange dot is a spoof type tablet data but annotated as live, which is define as data with noisy label. Overfitting on the such samples would cause large performance gaps on the test set. (b) Based on $\sigma^{\text{L}}$ which is the standard deviation of the Gaussian distribution of purple dot, DPM-LQ ``gives up'' embedding its feature representation $\mu$ closer to either of the spoof type label distribution, which speeds up the model converges. Further, by preventing the model from overfitting to data with noisy label (orange dot), DPM-DQ enlarges the distance between the Live/Spoof class center, which enhances the model robustness. (c) The $\sigma^{\text{D}}$ of the hard case (black cross) should similar to adjacent normal live samples, hence only the noisy data which is away from the Live/Spoof class center indicates the $\sigma^{\text{D}}$.
}
% \vspace{-13pt}
\label{figure:DPM_explanation}
\end{figure*}

\subsection{DPM-DQ}
From our in-depth analysis on False Reject (FR) and False Accept (FA) samples in typical FAS datasets, we observe that a certain amount of samples in FR and FA are noisy data yet with high confidences. 
Conventionally, $\mu_i$ will be fed into $g_{\mathcal{C}}(\cdot)$ for the cross-entropy loss of the Live/Spoof label: $\mathcal{L}_{\mathcal{C}}$ as follows:
\begin{equation}
\label{equation:conventional_loss_live_spoof}
    \mathcal{L}_{\mathcal{C}} = \frac{1}{N}\sum_{i}^{N}-\log \frac{e^{\omega_{c_i}\mu_{i}}}{\sum_{\mathcal{C}}^{}e^{\omega_{\mathcal{C}}\mu_{i}}},
\end{equation}
where $\omega_{c_i}$ is a row of vector of $\omega_\mathcal{C} \in \mathbb{R} ^{2\times B}$ which is the parameters of $g_{\mathcal{C}}(\cdot)$. 

\noindent \textbf{Distributional Modeling.} 
To correct the prediction confidence of noisy data, DPM-DQ formulates two Gaussian distributions in the latent space. Given an input image $x_i$ and its Live/Spoof ground truth $c_i$, $\omega_{c_i}$ can be regarded as its \textit{standard} feature representation which is the mean of the Gaussian distribution as follows: 
\begin{equation}
p(\mu_i \mid x_{i}) = \mathcal{N}(\mu_i;\omega_{c_i},({\sigma_{i}^{\text{D}}})^{2}\text{I}),
\end{equation}
where $\mu_i$ can be considered as a sample or the \textit{observed} feature representation in this distribution.
% with the variance $({\sigma_{i}^{\text{D}}})^{2}$.

\noindent \textbf{Live/Spoof Classification Loss.} 
For an observed feature representation, if we take the standard feature representation as its target, this Gaussian distribution of $\omega_{c_i}$ is formulated by maximizing the following log likelihood:
% \begin{equation}
% \label{equation:DPM-DQ_Gausian_distribution}
% \end{equation}
%
% Further, the  likelihood of Eq.~\ref{equation:DPM-DQ_Gausian_distribution} is formulated as follows:
\begin{equation}
\label{equation:data_uncertainty_1}
% \begin{aligned}
    % p(\mu_{i} \mid \omega_{y_{i}}) = \mathcal{N}(\mu_{i};\omega_{y_{i}},\sigma_{i}^{2}\text{I}) \\
    % p(\omega_{y_{i}} \mid \mu_{i}) = \frac{1}{\sqrt{2\pi{\sigma _{i}^{\text{d}}}^{2}}}exp(-\frac{(\omega_{y_{i}} - \mu_{i})^{2}}{2{\sigma _{i}^{\text{d}}}^{2}}) \\
     \log p(\omega_{c_i} \mid \mu_{i}) = -\frac{1}{2}\ln 2\pi - \frac{1}{2}\ln ({\sigma _{i}^{\text{D}}})^{2}-\frac{(\omega_{c_i} - \mu_{i})^{2}}{2({\sigma _{i}^{\text{D}}})^{2}},
% \end{aligned}
\end{equation}
where $({\sigma _{i}^{\text{D}}})^{2} = h_{\theta_{3}}(x_{i}) $, $\theta_{3}$ represents model parameters \textit{w.r.t} ${(\sigma _{i}^{\text{D}}})^{2}$. Different from Eq.~\ref{equation:conventional_loss_live_spoof}, $\mathcal{L}_{\mathcal{C}}$ in DPM-DQ is the reformulation of Eq.~\ref{equation:data_uncertainty_1} as the minimization of following loss function:
\begin{equation}
\label{equation:data_uncertainty_loss}
\begin{aligned}
\mathcal{L}_{c} & = \frac{1}{N}\sum_{N}^{}\frac{1}{2}(\ln ({\sigma _{i}^{\text{D}}})^{2}+\frac{(\omega_{\mathcal{C}_i} - \mu_{i})^{2}}{({\sigma _{i}^{\text{D}}})^{2}})+\frac{1}{2}\ln 2\pi.
% \\  
% & = \frac{1}{N}\sum_{N}^{}\frac{1}{2}f({\sigma _{i}^{\text{D}}}) + \frac{1}{2}\ln 2\pi, 
\end{aligned}
\end{equation} 
As will be discussed following in detail, after model converge, larger $(\sigma_i^\text{D})^2$ can be used as an indicator to noisy data, therefore in the inference phase, we leverage $
\exp~(-\frac{(\omega - \mu)^{2}}{2({\sigma^\text{D}})^{2} })$ to replace $\exp~(\omega\mu)$ in softmax. 
Therefore, for poor quality data, even if the model predicts high confidence for it, $({\sigma^\text{D}})^{2}$ can be used to correct its prediction confidence.

\noindent \textbf{Discussion.} 
In this discussion, we will explain: \textit{Why large $({\sigma^\text{D}})^{2}$ for noisy data}.
As Eq.~\ref{equation:data_uncertainty_loss} is optimized to minimum, we acquire that $(\sigma_i^\text{D})^2 = (\omega_{\mathcal{C}_i} - \mu_{i})^2$.
That is, $(\sigma_i^\text{D})^2$ tends to be proportional to the value of $(\omega_{\mathcal{C}_i} - \mu_{i})^2$ through optimizing. 
Among the samples of training data, $(\omega_{\mathcal{C}_i} - \mu_{i})^2$ of FA/FR and the noisy data are relatively large, $\sigma_i^\text{D}$ of both FA/FR and noisy data should be seemly large. 
However, since the feature representations $\mu$ among the adjacent distribution in the latent space are similar and $\sigma^{\text{D}}$ is predicated based on $\mu$, the $\sigma^{\text{D}}$ among the adjacent distribution should also be similar. 
As shown in dash circle of Fig.~\ref{figure:DPM_explanation} (c), the $\sigma^{\text{D}}$ of hard case (black cross) should be similar to normal samples (orange dots) around it. 
Therefore, the $\sigma^{\text{D}}$ of hard cases should not be so large, and only the noisy data which is away from the Live/Spoof class center indicates the larger $\sigma^{\text{D}}$, as shown in the right corner of Fig.~\ref{figure:DPM_explanation} (c).
% As shown in Fig.~\ref{figure:DPM_explanation} (c), 
% %
% both data noise (black dot) and hard case (black cross) are False Accept which away from the spoof class center. 
% From Eq.~\ref{equation:data_uncertainty_loss}, $(\sigma^\text{D})^2$ of both data noise and hard case should be large.  
% Therefore, we need to fix parameters of the feature extractor $h_{\theta_{1}}(\cdot)$ before DPM-DQ.
% %
% Since features embedding are determined, a small number of hard cases cannot make the $(\sigma^\text{D})^2$ of features embedding located around the adjacent latent space larger,
% %
% the larger $({\sigma^\text{D}})^{2}$ only belongs to noisy data which away from both the live class and the spoof class (as shown in Fig~\ref{figure:DPM_explanation}, noisy data locates in the bottom right corner).
% 

% \vspace{-5pt}
\subsection{Generalized DPM}
\label{sec:generalized co-uncertainty}
% \vspace{-2pt}
To deploy DPM in most of academic FAS datasets like Oulu-NPU~\cite{oulu-npu} and SiW~\cite{lyjauxuliary},
furthermore, to deploy DPM in practical uses, 
we propose a generalized version of DPM to tackle real-world FAS without the need of semantic annotations. 
Specifically, we define FAS datasets with sufficient semantic information as $\mathcal{D}^{\text{suf}}$, 
FAS datasets with deficient semantic information as $\mathcal{D}^{\text{def}}$.
\textbf{1)} Firstly, we train model $h_{\phi}(x)$ with $\mathcal{D}^{\text{suf}}$ to make $h_{\phi}(x)$ can tag semantic labels on $\mathcal{D}^{\text{def}}$. 
\textbf{2)} Secondly, for a semantic information $\mathcal{S}_k$ which is annotated in $\mathcal{D}^{\text{suf}}$ but not in $\mathcal{D}^{\text{def}}$, 
we leverage $h_{\phi}(x)$ to predict corresponding self-distributed semantic label $\mathcal{S}_k^\text{s}$ with $\mathcal{D}^{\text{def}}$. 
So far, for $\mathcal{D}^{\text{suf}}$, it contains Live/Spoof label $\mathcal{C}$, annotated semantic labels $\mathcal{S}^\text{a}$ which are already tagged in $\mathcal{D}^{\text{suf}}$, and self-distributed semantic labels $\mathcal{S}^\text{s}$.
% \footnote{We use $\mathcal{S}^\text{a}$ to represent all annotated semantic labels
% , $\mathcal{S}^\text{s}$ to represent all self-distributed semantic labels.}. 
%
\textbf{3)} Thirdly, leveraging DPM-LQ, we can make full use of the auxiliary role of both $\mathcal{S}^\text{a}$ and $\mathcal{S}^\text{s}$, even though inaccurate labels are inevitable in $\mathcal{S}^\text{s}$. 
%
% \textbf{4)} Fourthly, we need to finetune the Live/Spoof classification through $l_2$ feature normalization or ArcFace~\cite{deng2018arcface}\footnote{Because of the space limit, for detail explanation, please refer to \textit{Appendix}.}. 
%
\textbf{4)} Finally, DPM-DQ corrects the prediction confidence of noisy data.  
%
% Algorithm description is clearly shown as follows.
% %

% \setlength{\tabcolsep}{5pt}
% \begin{table}[!ht]
% \centering
% \ra{1.3}
% \label{table:generalized dual probabilistic modeling}
% \resizebox{0.45\textwidth}{!}{%
% \begin{tabular}{l}
% \toprule
% \textbf{Algorithm} :  Generalized dual probabilistic modeling                                 \\ \midrule
% \textbf{input}:  ${D}^{suf}$ and  $\mathcal{D}^{def}$ \\
% \textbf{output}: model $h_{\theta}(x)$                                \\
% \textbf{1 :} Training model $h_{\phi}(x)$ with  $\mathcal{D}_{suf}$ in multi-task fashion;  \\
% %
% \textbf{2 :} \textbf{for} semantic label $\mathcal{S}^\text{k}$ not in $\mathcal{D}^{def}$ \textbf{do} \\
% %
% \textbf{3 :} \hspace{10pt}    $\mathcal{S}_{\texttt{s}}^{\text{k}}$ = $h_{\phi}(\mathcal{D}_{suf})$. \\
% %
% \textbf{4 :} \textbf{end} \\
% %
% \textbf{5 :} Based on $\mathcal{S}_{\texttt{a}}$, $\mathcal{S}_{\texttt{s}}$ and $\mathcal{C}$,  \\ 
% \hspace{13pt}Training model $h_{\theta}(x)$ on $\mathcal{D}^{def}$ through coarse label aware learning. \\
% %
% \textbf{6 :} Finetuneing model $h_{\theta}(x)$ with $\mathcal{D}^{def}$ \\
% %
% \textbf{7 :} Fixing feature embedding network of $h_{\theta}(x)$. \\
% %
% \textbf{8 :} Modeling data quality of $\mathcal{D}^{def}$ based on $h_{\theta}(x)$ through data quality aware learning.

% \\ \bottomrule
% \end{tabular}%
% }
% \end{table}

\section{Experiments}

\begin{figure*}[t]
\centering
\includegraphics[width=\textwidth]{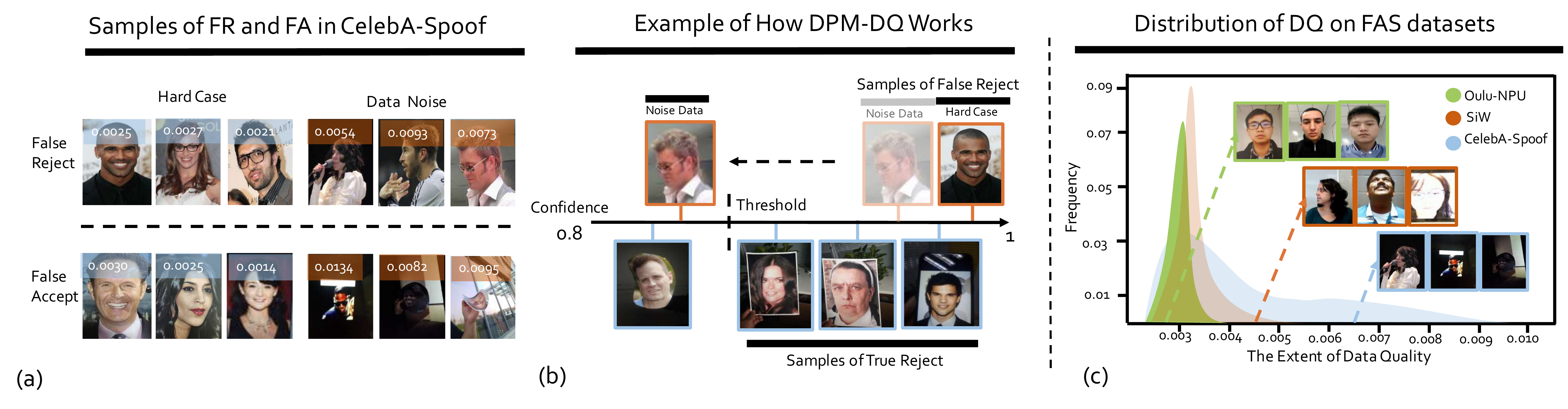}
% \vspace{-15pt}
\caption{(a) Both hard cases and noisy data appear in False Accept/False Reject in CelebA-Spoof. Values of $({\sigma^{\text{D}}})^{2}$ are shown on each image. $({\sigma^{\text{D}}})^{2}$ of data noise is higher than hard cases. (b) Based on the modeled data quality, the prediction confidence of noisy data decreases, even smaller than data which original confidence is smaller. (c) The distribution of the data quality on different datasets. Noted that the estimated data quality is proportional to the complexity of datasets. Specifically, since CelebA-Spoof is the highest diversity FAS dataset, the average value of the data quality of it is poorer than others. \textbf{Best viewed in color}.}
\vspace{-13pt}
\label{figure:data_quality_in_one_compressed}
\end{figure*}

In this section, we first evaluate the proposed method on standard FAS benchmarks. Then we explore how DPM can ease the impact of noise in FAS datasets through qualitative analyses. Last we carefully devise several ablation studies in order to shed more insights into generalized DPM.

\subsection{Datasets and Metrics}
\noindent \textbf{Datasets.}
Oulu-NPU~\cite{oulu-npu}, SiW~\cite{lyjauxuliary}, CASIA-MFSD~\cite{CASIA-MFSD}, Replay-Attack~\cite{Replay-Attack} and CelebA-Spoof~\cite{CelebA-Spoof} are used in our experiments. 
%
% Oulu-NPU, SiW, and CelebA-Spoof are high resolution datasets which proposed in recent years.  
% %
% Each dataset has several protocols that can be used to evaluate the FAS approach. 
% %
% CASIA-MFSD is used as the test set in the cross test setting.
% %
Specifically, CelebA-Spoof is a large scale dataset with sufficient semantic labels. However, since its large diversity and rich annotations, noisy data and inaccurate labels are a severe problem which hinders researchers to further explore it, we are the first to propose a network which can ease both label noise and data noise for CelebA-Spoof. Oulu-NPU and SiW are medium-scale datasets that lack enough semantic labels yet contain a certain degree of data noise. CASIA-MFSD and Replay-Attack are datasets which contain low resolution videos, which are used for cross-testing.

\noindent \textbf{Metrics.}
In SiW and Oulu-NPU, we follow original metrics and protocols to evaluate our model for the fair comparison, including APCER, BPCER, and ACER.
%
% \textit{i.e.}~Attack Presentation Classification Error Rate (APCER), 
% %
% Bona Fide Presentation Classification Error Rate (BPCER) and ACER. 
%
Except for three metrics above, 
we also leverage TPR@FPR to test our model in CelebA-Spoof. Further, 
Half Total Error Rate (HTER) is adopted in the cross testing.

\subsection{Implementation Details}
% The backbone of DPM is AENet$_{\mathcal{C}, \mathcal{S}, \mathcal{G}}$~\cite{CelebA-Spoof}.
The backbone of DPM is ResNet-18~\cite{ResNet}. 
Except for the Live/Spoof class, we leverage on the spoof type as semantic label for Oulu-NPU and SiW, and we leverage on spoof type, illumination conditions and face attributes as semantic labels for CelebA-Spoof.
% Specifically, the spoof type and illumination conditions are model probabilistic in CelebA-Spoof 
DPM-LQ is trained with Adam optimizer~\cite{adam} and the learning rate is $\text{10}^\text{-4}$ for 50 epochs. 
%
% However, Only spoof type is model probabilistic in Oulu-NPU and SiW. 
%
% Before DPM-DQ, ArcFace~\cite{deng2018arcface} is used to finetune the Live/Spoof classifier based on the SGD optimizer for 10 epochs. The learning rate is set to 1e-4.
Finally, the SGD optimizer is adopted for the training of DPM-DQ. The learning rate is set to $\text{10}^\text{-1}$ for 50 epochs.

\setlength{\tabcolsep}{5pt}
\begin{table}[t]
\footnotesize
\centering
\ra{1.1}
\caption{Results of the intra dataset test on Oulu-NPU. \textbf{Bolds} are the best results.}
\vspace{3pt}
\label{table:result of intra test on oulu-npu}
% \resizebox{0.48\textwidth}{!}{%
\begin{tabular}{@{}ccccc@{}}
\Xhline{1pt}
Prot.              & Methods        & APCER(\%)$\downarrow$ & BPCER(\%)$\downarrow$ & ACER(\%)$\downarrow$         \\ \hline
\multirow{8}{*}{1} & GRADIANT~\cite{Gradient}     & 1.3       & 12.5      & 6.9              \\
                   & BASN~\cite{KimBASN}          & 1.5       & 5.8       & 3.6              \\
                   & STASN~\cite{STASN}          & 1.2       & 2.5       & 1.9              \\
                   & Auxiliary~\cite{lyjauxuliary}     & 1.6       & 1.6       & 1.6              \\
                   & FaceDs~\cite{eccv18jourabloo}        & 1.2       & 1.7       & 1.5              \\
                   & Distangled~\cite{disentangled}        & 1.7       & 0.8       & 1.3              \\
                   & FAS-SGTD~\cite{wang2020FAS-SGTD}         & 2.0       & 0.0       & 1.0              \\
                    & CDCN~\cite{yztCDC}           & 0.4       & 1.7       & 1.0              \\
                   & BCN~\cite{yztHMP}           & 0.0         & 1.6       & 0.8              \\
                   & \textbf{Ours} & 0.7       & 0.6       & \textbf{0.6}     \\ \hline
\multirow{8}{*}{2} & FaceDs~\cite{eccv18jourabloo}        & 4.2       & 4.4       & 4.3              \\
                   & Auxiliary~\cite{lyjauxuliary}     & 2.7       & 2.7       & 2.7              \\
                   & BASN~\cite{KimBASN}          & 2.4       & 3.1       & 2.7              \\
                   & GRADIANT~\cite{Gradient}      & 3.1       & 1.9       & 2.5              \\
                   & Distangled~\cite{disentangled}        & 1.1       & 3.6       & 2.4               \\
                   & STASN~\cite{STASN}          & 4.2      & 0.3       & 2.2              \\
                   & FAS-SGTD~\cite{wang2020FAS-SGTD}         & 2.5       & 1.3       & 1.9              \\
                   & BCN~\cite{yztHMP}           & 2.6       & 0.8       & 1.7              \\
                   & CDCN~\cite{yztCDC}           & 1.5       & 1.4       & 1.5              \\
                   & \textbf{Ours} & 2.4       & 0.4       & \textbf{1.4}     \\ \hline
\multirow{8}{*}{3} & GRADIANT~\cite{Gradient}     & 2.6$\pm $ 3.9   & 5.0$\pm $ 5.3   & 3.8$\pm $ 2.4          \\
                   & BASN~\cite{KimBASN}          & 1.8$\pm $ 1.1   & 3.5$\pm $ 3.5   & 2.7$\pm $ 1.6          \\
                   & FaceDS~\cite{eccv18jourabloo}        & 4.0$\pm $ 1.8   & 3.8$\pm $ 1.2   & 3.6$\pm $ 1.6          \\
                   & Auxuliary~\cite{lyjauxuliary}     & 2.7$\pm $ 1.3   & 3.1$\pm $ 1.7   & 2.9$\pm $ 1.5          \\
                   & STASN~\cite{STASN}          & 4.7 $\pm $ 3.9   & 0.9 $\pm $ 1.2   & 2.8 $\pm $ 1.6              \\
                   & FAS-SGTD~\cite{wang2020FAS-SGTD}         & 3.2$\pm $ 2.0   & 2.2$\pm $ 1.4   & 2.7$\pm $ 0.6          \\
                   & Distangled~\cite{disentangled}      & 2.8$\pm $ 2.2   & 1.7 $\pm $ 2.6   & 2.2$\pm $ 2.2 \\
                   & BCN~\cite{yztHMP}           & 2.8$\pm $ 2.4   & 2.3$\pm $ 2.8   & 2.5$\pm $ 1.1          \\
                   & CDCN~\cite{yztCDC}          & 2.4$\pm $ 1.3   & 2.2$\pm $ 2.0   & 2.3$\pm $ 1.4          \\
                   & \textbf{Ours} & 1.9$\pm $ 1.8   & 1.4$\pm $ 1.4   & \textbf{1.6$\pm $ 1.5} \\ \hline
\multirow{8}{*}{4} & GRADIANT~\cite{Gradient}      & 5.0$\pm $ 4.5   & 15.0$\pm $ 7.1  & 10.0$\pm $ 5.0         \\
                   & Auxiliary~\cite{lyjauxuliary}     & 9.3$\pm $ 5.6   & 10.4$\pm $ 6.0  & 9.5$\pm $ 6.0          \\  
                   & STASN~\cite{STASN}          & 6.7 $\pm $ 10.6   & 8.3 $\pm $ 8.4   & 7.5 $\pm $ 4.7              \\
                    & CDCN~\cite{yztCDC}           & 4.6$\pm $ 4.6   & 9.2$\pm $ 8.0   & 6.9$\pm $ 2.9          \\
                   & FaceDS~\cite{eccv18jourabloo}        & 1.2$\pm $ 6.3   & 6.1$\pm $ 5.11  & 5.6$\pm $ 5.7          \\
                   & BASN~\cite{KimBASN}          & 6.4$\pm $ 8.6   & 7.5$\pm $ 6.9   & 5.2$\pm $ 3.7          \\
                   & BCN~\cite{yztHMP}           & 2.9$\pm $ 4.0   & 7.5$\pm $ 6.9   & 5.2$\pm $ 3.7          \\
                   & FAS-SGTD~\cite{wang2020FAS-SGTD}         & 6.7$\pm $ 7.5   & 3.3$\pm $ 4.1   & 5.0$\pm $ 2.2          \\
                   & Distangled~\cite{disentangled}      & 5.4$\pm $ 2.9   & 3.3 $\pm $ 6.0   & 4.4 $\pm $ 3.0 \\

                   & \textbf{Ours} & 3.9$\pm $ 6.0   & 1.1$\pm $ 1.3   & \textbf{2.4$\pm $ 4.4} \\ \Xhline{1pt}

\end{tabular}%
% }
% \vspace{-8pt}
\end{table}

\setlength{\tabcolsep}{5pt}
\begin{table}[t]
% \tiny
\centering
\ra{1.0}
\caption{Results of the intra-dataset test on CelebA-Spoof. \textbf{Bolds} are the best results.}
\vspace{3pt}
\label{table:the results of intra-dataset test on CelebA-Spoof}
\resizebox{0.48\textwidth}{!}{%
\begin{tabular}{@{}ccccccc@{}}
\Xhline{1pt}
\multirow{2}{*}{Methods} &
  \multicolumn{3}{c}{TPR (\%)$\uparrow$ } &

  \multirow{2}{*}{APCER (\%)$\downarrow$ } &
  \multirow{2}{*}{BPCER (\%)$\downarrow$ } &
  \multirow{2}{*}{ACER (\%)$\downarrow$ } \\ \cline{2-4}
   & FPR = 1\% & FPR = 0.5\%   & FPR = 0.1\%   &        &       &           \\ \hline
Auxiliary~\cite{CelebA-Spoof} &
97.3      & 95.2          & 83.2          & 5.71 & 1.41 & 3.56 \\ 
BASN~\cite{CelebA-Spoof}   &
  98.9 &
  97.8 &
  \textbf{90.9} &
  4.0 &
  1.1 &
  2.6 \\ %\hline
AENet$_{\mathcal{C},\mathcal{S}, \mathcal{G}}$~\cite{CelebA-Spoof}       & 
98.9 & 
97.3          & 
87.3          & 
2.29          & 
0.96         & 
1.63          \\
\textbf{Ours} & 
\textbf{99.3} & 
\textbf{98.2} & 
89.28          & 
\textbf{0.84}  & 
\textbf{0.82}  & 
\textbf{0.83}          \\   \Xhline{1pt} %\hline
\end{tabular}%
}
% \vspace{-3pt}
\end{table}

\setlength{\tabcolsep}{5pt}
\begin{table}[t]
% \tiny
\centering
\ra{1.1}
\caption{Results of the cross-domain test on CelebA-Spoof. \textbf{Bolds} are the best results.}
\vspace{3pt}
\label{table:the results of cross-domain test on CelebA-Spoof}
\resizebox{0.48\textwidth}{!}{%
\begin{tabular}{cccccccccc}
\Xhline{1pt}
\multirow{2}{*}{Prot.} &
  \multirow{2}{*}{Methods} &
  \multicolumn{3}{c}{TPR (\%) $\uparrow$} &
  \multirow{2}{*}{APCER (\%)$\downarrow$} &
  \multirow{2}{*}{BPCER (\%)$\downarrow$} &
  \multirow{2}{*}{ACER (\%)$\downarrow$} \\ \cline{3-5}
 &
   &
  FPR = 1\% &
  FPR = 0.5\% &
  FPR = 0.1\% &
   &
   &
   \\ \hline
   
\multirow{3}{*}{1} &
AENet$_{\mathcal{C}, \mathcal{S}, \mathcal{G}}$~\cite{CelebA-Spoof} &
  95.0 &
  91.4 &
  73.6 &
  4.09 &
  2.09 &
  3.09  \\ 

& \textbf{Ours}&
  \textbf{96.7} &
  \textbf{93.9} &
  \textbf{79.0} &
  \textbf{2.09} &
  \textbf{1.57} &
  \textbf{1.83}  \\  \hline

\multirow{3}{*}{2} &
AENet$_{\mathcal{C}, \mathcal{S}, \mathcal{G}}$~\cite{CelebA-Spoof} &
  
  \# &
  \# &
  \# &
  4.94$\pm $3.42 &
  1.24$\pm $0.73 &
  3.09$\pm $2.08  \\
  
& \textbf{Ours} &
  \# &
  \# &
  \# &
  \textbf{1.02$\pm $0.29} &
  \textbf{0.72$\pm $0.62} &
  \textbf{0.73$\pm $0.64}  \\
  \Xhline{1pt}
\end{tabular}%
}
% \vspace{-15pt}
\end{table}

\setlength{\tabcolsep}{5pt}
\begin{table}[t]
\scriptsize
\centering
\caption{Cross-dataset test results of CelebA-Spoof. \textbf{Bolds} are the best results.}
\vspace{3pt}
\label{table:The Cross-dataset test results of CelebA-Spoof.}
\centering
% \resizebox{0.48\textwidth}{!}{%
\begin{tabular}{@{}cccc@{}}
\Xhline{1pt}
Methods    &  Training     & Test     & HTER (\%) $\downarrow$      \\ \hline
AENet$_{\mathcal{C},\mathcal{S}, \mathcal{G}}$~\cite{CelebA-Spoof} &
CelebA-Spoof & CASIA-MFSD & 11.9 \\
\textbf{Ours}&
CelebA-Spoof & CASIA-MFSD & \textbf{10.0} \\
\Xhline{1pt}
\end{tabular}%
% }
% \vspace{-8pt}
\end{table}

\setlength{\tabcolsep}{5pt}
\begin{table}[t]
\small
\centering
% 
%\fontsize{10pt}{10pt}
%\scriptsize
\ra{1.1}
\caption{The results of cross-dataset testing between CASIA-MFSD and Replay-Attack. The evaluation metric is HTER(\%). \textbf{Bolds} are the best results.}
\vspace{3pt}
\label{table:the results of cross-dataset testing between CASIA-MFSD and Replay-Attack}
% \resizebox{0.48\textwidth}{!}{%
\begin{tabular}{@{}ccc@{}}
\Xhline{1pt}
Train & CASIA-MFSD & Replay-Attack \\ %\midrule
Test  & Replay-Attack & CASIA-MFSD       \\ \hline
Motion-Mag~\cite{motionmag} &
50.1\% & 47.0\% \\ 
Spectral cubes~\cite{spectralcubes} &
34.4\% & 50.0\% \\ 
LowPower~\cite{lowPower} &
30.1\% & 35.6\% \\ 
CNN~\cite{CNN} &
48.5\% & 45.5\% \\ 
STASN~\cite{STASN} &
31.5\% & 30.9\% \\ 
FaceDs~\cite{eccv18jourabloo} &
28.5\% & 41.1\% \\ 
Auxuliary~\cite{lyjauxuliary} &
27.6\% & \textbf{28.4} \% \\ 
BASN~\cite{KimBASN} &
23.6\% & 29.9\% \\
BCN~\cite{yztHMP} &
16.6 \% & 36.4\% \\
Distangled~\cite{disentangled} &
22.4\% & 30.3\% \\
CDCN~\cite{yztCDC} &
\textbf{15.5} \% & 32.6\% \\   
\textbf{Ours} &
20.4 \% & 35.2\% \\  \Xhline{1pt} %\hline
\end{tabular}%
% }
% \vspace{-10pt}
\end{table}

\setlength{\tabcolsep}{5pt}
\begin{table}[t]
\footnotesize
\centering
\ra{1.3}
\caption{Results of intra dataset test on SiW. \textbf{Bolds} are the best results.}
\vspace{3pt}
\label{table:result of intra test on siw}
% \resizebox{0.4\textwidth}{!}{%
\begin{tabular}{@{}ccccc@{}}
\Xhline{1pt}
Prot.              & Methods        & APCER(\%)$\downarrow$ & BPCER(\%)$\downarrow$ & ACER(\%)$\downarrow$         \\ \hline
\multirow{6}{*}{1} & Auxiliary~\cite{lyjauxuliary}      & 3.58       & 3.58      & 3.58           \\
                  & FAS-SGTD~\cite{wang2020FAS-SGTD}         & 0.64       & 0.17    &   0.40 		\\
                  & BASN~\cite{KimBASN}           & -         & -       & 0.37              \\
                  & BCN~\cite{yztHMP}           & 0.55        & 0.17       & 0.36              \\
                  & \textbf{Ours} & 0.46       & 0.19       & 0.28     \\
                  & CDCN~\cite{yztCDC}           & 0.07       & 0.17       & \textbf{0.12}    \\ \hline
\multirow{6}{*}{2} & Auxiliary~\cite{lyjauxuliary}     & 0.57 $\pm $ 0.60  &  0.57 $\pm $ 0.60 & 0.57 $\pm $ 0.60                           \\

& BASN~\cite{KimBASN}          & -       & -      & 0.12 $\pm $ 0.03 \\

& \textbf{Ours} & 0.00 $\pm$ 0.00 &	0.26 $\pm$0.20 & 0.13 $\pm$0.10 \\

& BCN~\cite{yztHMP}   & 0.08 $\pm $ 0.17 &	1.15 $\pm$ 0.00	& 0.11 $\pm $ 0.08              \\

& CDCN~\cite{yztCDC}	& 0.00 $\pm $ 0.00 & 0.13 $\pm $ 0.09 &	0.06 $\pm $ 0.04     \\ 

 & FAS-SGTD~\cite{wang2020FAS-SGTD} &	0.00 $\pm $ 0.00	& 0.04 $\pm $ 0.08 & 	\textbf{0.02 $\pm $ 0.04} \\\hline
\multirow{6}{*}{3} & Auxiliary~\cite{lyjauxuliary}      & 8.31 $\pm $ 3.81   &  8.31 $\pm $ 3.81    & 8.31 $\pm $ 3.81          \\
& BASN~\cite{KimBASN}       & -   & -   & 6.45 $\pm $ 1.80          \\
 & FAS-SGTD~\cite{wang2020FAS-SGTD}	&  2.63 $\pm $ 3.72 &	2.92 $\pm $ 3.42 &	2.78 $\pm $ 3.57 \\ 
& BCN~\cite{yztHMP}           & 2.55 $\pm $ 0.89   &  2.34 $\pm $ 0.47   & 2.45 $\pm $ 0.68          \\
 & \textbf{Ours} & 2.23 $\pm $ 4.8 &	2.53 $\pm $ 4.2 & 2.44 $\pm $ 4.50 \\
& CDCN~\cite{yztCDC}    &       1.67 $\pm $ 0.11	 & 1.76 $\pm $ 0.12& \textbf{1.71 $\pm $ 0.10}        \\ \Xhline{1pt}
\end{tabular}%
% }
\end{table}

\subsection{Comparison with State-of-The-Arts}
\noindent \textbf{Intra Dataset Test.}
The intra dataset test is carried out on Oulu-NPU, SiW, and CelebA-Spoof.
For Oulu-NPU and SiW, four protocols and three protocols are designed respectively to evaluate the generalization capability of FAS methods. 
For CelebA-Spoof, the intra dataset test includes intra-dataset test and cross-domain test. 
Intra-dataset test is designed to evaluate the overall capability of the proposed method. 
Two protocols in the cross-domain test are designed to evaluate the performance of FAS methods under controlled domain shifts. 
As shown in Table.~\ref{table:result of intra test on oulu-npu}, DPM ranks the first on all 4 protocols of Oulu-NPU, 
which indicates the great generalization ability of our method on different environment conditions, spoof types, and input sensors. 
Besides, as shown in Table.~\ref{table:the results of intra-dataset test on CelebA-Spoof} and Table~\ref{table:the results of cross-domain test on CelebA-Spoof}, %
DPM achieves the best performance among both the intra-dataset test and the cross-domain test of CelebA-Spoof, which indicates an excellent capacity of DPM on a large scale dataset.
Specifically, compared to AENet$_{\mathcal{C}, \mathcal{S}, \mathcal{G}}$, 
our method improves 40.8\% and 73.4\% respectively in Table.~\ref{table:the results of cross-domain test on CelebA-Spoof}.
In addition, our method achieves comparable results in SiW as shown in Table~\ref{table:result of intra test on siw}.
% Table~\ref{table:result of intra test on siw}.

\noindent \textbf{Cross Dataset Test.}
We conduct experiments to evaluate the generalization ability of methods. 
As shown in Table.~\ref{table:The Cross-dataset test results of CelebA-Spoof.}, 
DPM outperform prior state-of-the-art results on the cross-dataset setting of CelebA-Spoof, which indicates great generalization ability of DPM. Besides, DPM achieves comparable results on the cross-dataset testing between CASIA-MFSD and Replay-Attack as shown in Table.~\ref{table:the results of cross-dataset testing between CASIA-MFSD and Replay-Attack}.

\noindent \textbf{Evaluation on Different Backbone.}
To evaluate DPM comprehensively, we conduct DPM with heavier model-Xception~\cite{xception} on three benchmarks of CelebA-Spoof. As shown in the Table.~\ref{table:the results of intra-dataset test on CelebA-Spoof xception}, Table.~\ref{table:the results of cross-domain test on CelebA-Spoof xception} and Table.~\ref{table:The Cross-dataset test results of CelebA-Spoof xception}, DPM achieves state-of-the-arts results on these benchmarks, which indicates that DPM can be incorporated into existing deep networks seamlessly and efficiently.

\setlength{\tabcolsep}{5pt}
\begin{table}[t]
% \tiny
\centering
\ra{1.1}
\caption{Results of the cross-domain test on CelebA-Spoof. \textbf{Bolds} are the best results, all methods are based on Xception.}
\vspace{3pt}
\label{table:the results of cross-domain test on CelebA-Spoof xception}
\resizebox{0.48\textwidth}{!}{%
\begin{tabular}{cccccccccc}
\Xhline{1pt}
\multirow{2}{*}{Prot.} &
  \multirow{2}{*}{Methods} &
  \multicolumn{3}{c}{TPR (\%) $\uparrow$} &
  \multirow{2}{*}{APCER (\%)$\downarrow$} &
  \multirow{2}{*}{BPCER (\%)$\downarrow$} &
  \multirow{2}{*}{ACER (\%)$\downarrow$} \\ \cline{3-5}
 &
   &
  FPR = 1\% &
  FPR = 0.5\% &
  FPR = 0.1\% &
   &
   &
   \\ \hline
   
\multirow{3}{*}{1} &
AENet$_{\mathcal{C}, \mathcal{S}, \mathcal{G}}$~\cite{CelebA-Spoof} &
  96.9 &
  93.0 &
  83.5 &
  3.00 &
  1.48 &
  2.24  \\ 

& \textbf{Ours}&
  \textbf{97.2} &
  \textbf{95.4} &
  \textbf{87.2} &
  \textbf{2.63} &
  \textbf{0.62} &
  \textbf{1.63}  \\  \hline

\multirow{3}{*}{2} &
AENet$_{\mathcal{C}, \mathcal{S}, \mathcal{G}}$~\cite{CelebA-Spoof} &
  
  \# &
  \# &
  \# &
  4.77$\pm $4.12 &
  1.23$\pm $1.06 &
  3.00$\pm $2.90  \\
  
& \textbf{Ours} &
  \# &
  \# &
  \# &
  \textbf{1.18$\pm$0.72} &
  \textbf{0.64$\pm $0.17} &
  \textbf{0.91$\pm$0.45}  \\
  \Xhline{1pt}
\end{tabular}%
}
% \vspace{-15pt}
\end{table}

\setlength{\tabcolsep}{5pt}
\begin{table}[t]
\scriptsize
\centering
\caption{Cross-dataset test results of CelebA-Spoof. \textbf{Bolds} are the best results, all methods are based on Xception.}
\vspace{3pt}
\label{table:The Cross-dataset test results of CelebA-Spoof xception}
\centering
% \resizebox{0.48\textwidth}{!}{%
\begin{tabular}{@{}cccc@{}}
\Xhline{1pt}
Methods    &  Training     & Test     & HTER (\%) $\downarrow$      \\ \hline
AENet$_{\mathcal{C},\mathcal{S}, \mathcal{G}}$~\cite{CelebA-Spoof} &
CelebA-Spoof & CASIA-MFSD & 13.1 \\
\textbf{Ours}&
CelebA-Spoof & CASIA-MFSD & \textbf{11.7} \\
\Xhline{1pt}
\end{tabular}%
% }
% \vspace{-8pt}
\end{table}

% \setlength{\tabcolsep}{5pt}
% \begin{table}[t]
% \centering
% \ra{1.3}
% \caption{Experiments of finetune before DPM-DQ on CelebA-Spoof intra-dataset test. It indicates finetune plays an important role in DPM. The performance of DPM w/o FT on CelebA-Spoof become worse than DPM. FT indicates finetune.}
% \label{table:the results of finetune CelebA-Spoof}
% \resizebox{0.45\textwidth}{!}{%
% \begin{tabular}{@{}cccc@{}}
% \toprule 
% Method & 
% APCER (\%)$\downarrow$  &
% BPCER (\%)$\downarrow$  &
% ACER (\%)$\downarrow$  \\ \midrule
% AENet$_{\mathcal{C}, \mathcal{S}, \mathcal{G}}$~\cite{CelebA-Spoof} &
% 2.29          & 
% 0.96         & 
% 1.63          \\
% AENet$_{\mathcal{C}, \mathcal{S}, \mathcal{G}}$ w/ FT  &
% 1.99          & 
% 0.93          & 
% 1.46   \\
% DPM w/o FT   &
% 1.25          & 
% 0.89         & 
% 1.08   \\
% \textbf{DPM w/ FT} & 
% \textbf{0.77}  & 
% \textbf{0.82}  & 
% \textbf{0.76}          \\                \bottomrule %\hline
% \end{tabular}%
% }
% \end{table}
%
% \begin{figure}[t]
% \centering
% \includegraphics[width=0.4\textwidth]{Dual_Probabilistic_Modeling/figure/data_quality_in_one_compressed.pdf}
% \caption{(a) Examples of FAR and FRR in CelebA-Spoof. Both hard case and noise data appear. values of $({\sigma^{\text{D}}})^{2}$ are shown on each image.
% }
% \label{figure:FARandFRR}
% \end{figure}

%data_quality_in_one_compressed
\subsection{Further Analysis}
In this part, we conduct experiments to explain how and why DPM can ease the impact of noise in FAS datasets. Further we demonstrate that the relationship between DPM and quality assessment models (QAM). Specifically, all experiments in this part are conducted on CelebA-Spoof.

\noindent \textbf{The Working Mechanism of DPM-LQ.}
Fig.~\ref{figure:DPM_explanation} (a) presents the distribution of the features representation in the latent space predicted by the model trained with the Live/Spoof label and auxiliary semantic labels.
%AENet$_{\mathcal{C},\mathcal{S}, \mathcal{G}}$.
%
The classification boundary of three spoof types (PC, phone, table) is unclear, which indicates label ambiguous. 
Specifically, for the purple dot which closes to any of the spoof type class centers, pushing model embedding its feature representation close to the Phone class center but away from other class centers makes it hard to converge. 
Besides, label noise refers to the orange dot, which is a spoof data but annotated as live. 
Overfitting to such data with label noise is harmful for model generalization.
DPM-LQ eases the above problems as illustrated in Fig.~\ref{figure:DPM_explanation} (b): 
\textbf{1)} DPM-LQ chooses to ``give up'' the orange dot by leveraging $\sigma^{\text{L}}$, which is the standard deviation of its Gaussian distribution, to prevent the model from overfitting to the orange dot, DPM-LQ hence can effectively solve the noisy label problem. 
\textbf{2)} 
DPM-LQ leverages $z$ to replace $\mu$ for the calculation of semantic classification loss, which prevents the distribution of each ambiguous label from overly distinguishing from each other. As a result, DPM-LQ significantly speeds up the model converge.

\noindent \textbf{Understand Data Quality.} 
As shown in Fig.~\ref{figure:data_quality_in_one_compressed} (a), both noisy data and hard cases appear in False Reject (FR) and False Accept (FA). 
Data noise refers to images with low quality, such as a face with the large variation for live data or images under specific illumination conditions for spoof data.
Fortunately, since the extent of data noise $({\sigma^{\text{D}}})^{2}$ of noisy data is much larger than hard cases, as shown in Fig.~\ref{figure:data_quality_in_one_compressed} (b), the prediction confidence of noisy data decreases through correcting by $({\sigma^{\text{D}}})^{2}$, In this way, noisy data can be filtered out from FR, which further improves FRR. 
Besides, as shown in Fig.~\ref{figure:data_quality_in_one_compressed} (c), the ``variance'' of the data quality increases in the following order: Oulu-NPU $<$ SiW $<$ CelebA-Spoof. This order is proportional to the extent of the data diversity in different modern FAS datasets. Specifically, CelebA-Spoof is the richest diversity FAS dataset.

\noindent \textbf{Feature Normalization is Indispensable.}
Before DPM-DQ, we leverage the cross entropy loss to train our model, $\omega_{\mathcal{C}_i}\mu_{i}$ is part of the numerator of the cross entropy loss, which can be considered as ``distance'' between $\omega_{\mathcal{C}_i}$ and $\mu_{i}$.
$\sigma_i^\text{D}$ can be modeled as data quality based on the assumption that FR and FA are the samples, of which $(\omega_{\mathcal{C}_i} - \mu_{i})^2$ is large. $(\omega_{\mathcal{C}_i} - \mu_{i})^2$ can also be considered as ``distance'' between $\omega_{\mathcal{C}_i}$ and $\mu_{i}$ under the Gaussian distribution.
However, $\omega_{\mathcal{C}_i}\mu_{i}$ in the cross entropy loss is not necessarily proportion to $-(\omega_{\mathcal{C}_i} - \mu_{i})^2$ in the Gaussian distribution. Given we further fix parameters of $h_{\theta_{1}}(\cdot)$ during DPM-DQ, we need to reformulate $\omega_{\mathcal{C}_i}$ and $\mu_{i}$, so that $\omega_{\mathcal{C}_i} \mu_{i}$ is proportion to $-(\omega_{\mathcal{C}_i} - \mu_{i})^2$.
Therefore, we finetune $\omega_{\mathcal{C}_i}$ and $\mu_{i}$ based on $l2$ normalization,
$\left \| \omega_{\mathcal{C}_i} \right \|$ and $\left \| \mu_{i} \right \|$ are the normalized $\omega_{\mathcal{C}_i}$ and $\mu_{i}$ respectively. That is, $\left \| \omega_{\mathcal{C}_i} \right \|\left \| \mu_{i} \right \|$ in the cross entropy loss is proportional to $-(\omega_{\mathcal{C}_i} - \mu_{i})^2$ in the Gaussian distribution. 
Besides, as shown in Table~\ref{table:the results of finetune CelebA-Spoof}, the performance of DPM without finetune becomes worse. It indicates that finetune is extremely important to DPM-DQ.

\setlength{\tabcolsep}{5pt}
\begin{table}[t]
% \tiny
\centering
\ra{1.0}
\caption{Results of the intra-dataset test on CelebA-Spoof. \textbf{Bolds} are the best results, all methods are based on Xception.}
\vspace{3pt}
\label{table:the results of intra-dataset test on CelebA-Spoof xception}
\resizebox{0.48\textwidth}{!}{%
\begin{tabular}{@{}ccccccc@{}}
\Xhline{1pt}
\multirow{2}{*}{Methods} &
  \multicolumn{3}{c}{TPR (\%)$\uparrow$ } &

  \multirow{2}{*}{APCER (\%)$\downarrow$ } &
  \multirow{2}{*}{BPCER (\%)$\downarrow$ } &
  \multirow{2}{*}{ACER (\%)$\downarrow$ } \\ \cline{2-4}
   & FPR = 1\% & FPR = 0.5\%   & FPR = 0.1\%   &        &       &           \\ \hline
AENet$_{\mathcal{C},\mathcal{S}, \mathcal{G}}$~\cite{CelebA-Spoof}       & 
\textbf{99.2} & 
98.4          & 
94.2         & 
3.72          & 
0.82         & 
2.27          \\
\textbf{Ours} & 
99.0 & 
\textbf{98.5} & 
\textbf{94.7}         & 
\textbf{1.37}  & 
\textbf{0.62}  & 
\textbf{0.99}           \\   \Xhline{1pt} %\hline
\end{tabular}%
}
% \vspace{-3pt}
\end{table}

\setlength{\tabcolsep}{5pt}
\begin{table}[t]
\centering
\footnotesize
\ra{1.3}
\caption{Experiments of finetune before DPM-DQ on CelebA-Spoof intra-dataset test. It indicates finetune plays an important role in DPM. The performance of DPM w/o FT on CelebA-Spoof becomes worse than DPM. FT indicates finetune.}
\vspace{3pt}
\label{table:the results of finetune CelebA-Spoof}
% \resizebox{0.45\textwidth}{!}{%
\begin{tabular}{@{}cccc@{}}
\Xhline{1pt} 
Methods & 
APCER (\%)$\downarrow$  &
BPCER (\%)$\downarrow$  &
ACER (\%)$\downarrow$  \\ \hline
AENet$_{\mathcal{C}, \mathcal{S}, \mathcal{G}}$~\cite{CelebA-Spoof} &
2.29          & 
0.96         & 
1.63          \\
AENet$_{\mathcal{C}, \mathcal{S}, \mathcal{G}}$ w/ FT  &
1.99          & 
0.93          & 
1.46   \\
DPM w/o FT   &
1.25          & 
0.89         & 
1.08   \\
\textbf{DPM w/ FT} & 
\textbf{0.77}  & 
\textbf{0.82}  & 
\textbf{0.76}          \\  \Xhline{1pt}
\end{tabular}%
% }
\end{table}

\begin{figure*}[t]
\centering
\includegraphics[width=0.99\textwidth]{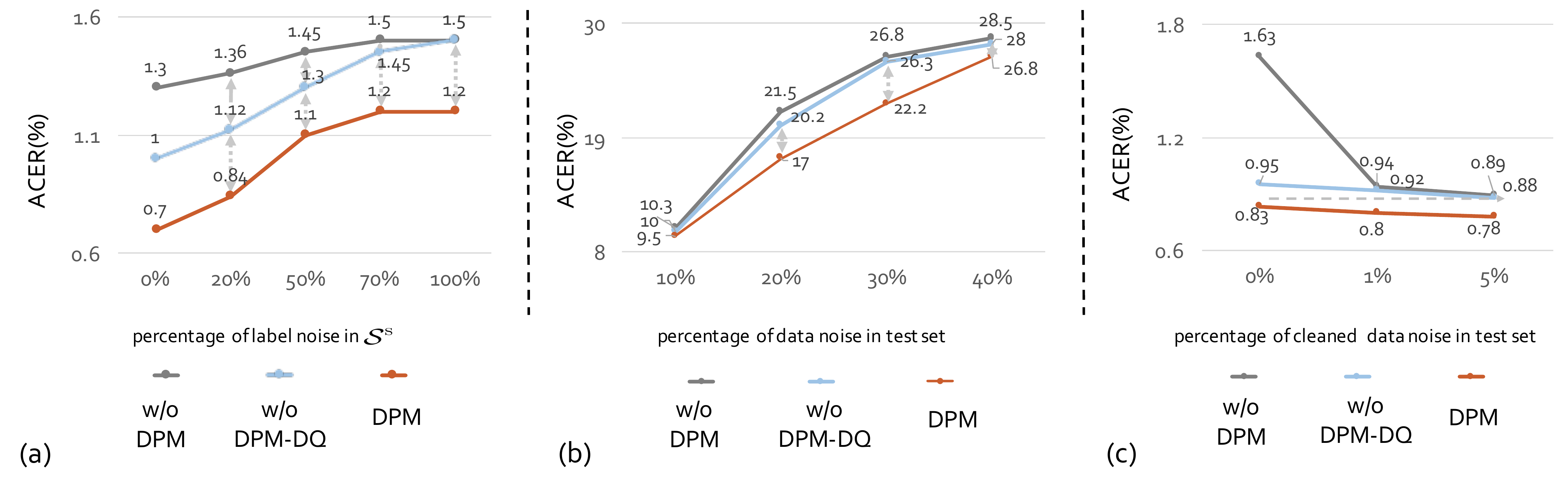}
% \vspace{-7pt}
\caption{(a) When the proportion of label noise becomes slightly bigger (less than 50\%), DPM-LQ can still improve the classification performance. However, as label noise becomes extremely serious, DPM-LQ can barely work. (b) When the proportion of data noise is low-level, DPM-DQ can \textit{slightly} improve model performance. Further, when the proportion of data noise is mid-level (20\%, 30\%), DPM-DQ can \textit{apparently} improve the performance of DPM. However, DPM-DQ barely works when data noise becomes extremely serious. (c) Only testing on the test set where top 5\% of relative low-quality data has been cleaned, the model without DPM can achieve the comparable performance as the model with DPM which is tested on the whole test set. Besides, DPM can further improve the performance of ``cleaner'' test data.
}
\label{figure:noise}
% \vspace{-8pt}
\end{figure*}

\noindent \textbf{The Relationship with Quality Assessment Model.} 
Quality assessment models (QAM)~\cite{best2018learning} are largely used to filter the noisy data. 
To discuss the relationship between DPM and QAM, we conduct several experiments based on the test set of the CelebA-Spoof intra-dataset test setting. 
% \footnote{This experiment conducts on CelebA-Spoof because the data in CelebA-Spoof is comparable diverse as shown in Figure~\ref{figure:data_quality_in_one_compressed} (c).}, 
%
Specifically, we compare the performance of AENet$_{\mathcal{C}, \mathcal{S}, \mathcal{G}}$ based on \textit{cleaner test set}, which cleaned by FaceQNet~\cite{faceqnet} with the performance of DPM based on \textit{whole test set}.~\footnote{\scriptsize
{FaceQNet the state-of-the-art QAM model trained with VGGFace2~\cite{VGGFace2}.}}
To forming different cleanliness test sets, we clean up the top 1\% to 5\% of relative low-quality data which is ranked by FaceQNet.
As shown in Fig.~\ref{figure:noise} (c), Only testing on the test set where the top 5\% of relative low-quality data has been cleaned, 
AENet$_{\mathcal{C}, \mathcal{S}, \mathcal{G}}$ can achieve comparable performance (ACER: 0.88) as the model with DPM tested on whole test set (ACER: 0.83). 
Therefore, comparing to FaceQNet, which costs extra overhead for training, DPM is a more efficient solution for cleaning data noise. 
Besides, as shown in Fig.~\ref{figure:noise} (c), DPM can further improve the model performance on a cleaner dataset, which indicates low-quality images ranked by FaceQNet and low-quality images ranked by DPM do not necessarily overlap, and there is a complementary relationship between them. 

\subsection{Ablation Study}
% In this part, 
% first, we explain the importance of finetune before DPM-DQ through the study on CelebA-Spoof intra-dataset test. 
%
Based on the $\mathcal{D}^{\text{def}}$ Oulu-NPU, we further explore the ability of generalized DPM pipeline. Specifically, we deliberately tag the self-distributed spoof type label on Oulu-NPU, even if such label has been handily annotated. Presumably, if the performance of DPM based on \textit{self-distributed spoof type label} is closed to the performance of DPM based on \textit{annotated spoof type label}, generalized DPM pipeline would prove to be an effective solution to tackle real-world FAS datasets which lacks handily annotated semantic labels.
Experiments are conducted on Oulu-NPU protocol 1.
%
% Notations used are listed in \textit{Appendix}.
% Table~\ref{table:notations}. 
%
Specifically, $\mathcal{D}^{\text{suf}}$ refers to CelebA-Spoof in this part. 
Baseline refers to the vanilla binary classification fashion in FAS.

\setlength{\tabcolsep}{5pt}
\begin{table}[t]
\small
\centering
\ra{1.1}
\caption{Several quantitative results of ablation studies. $\mathcal{S}^{\text{s}}$ refers to the self-distributed spoof type label and $\mathcal{S}^{\text{s}}$ refers to the annotated spoof type label. (a) DPM with $\mathcal{S}^{\text{s}}$ outperforms the baseline. It indicates the validity of the self-distribution annotations. (b) With the help of DPM-LQ, DPM with $\mathcal{S}^{\text{s}}$ achieves comparable results comparing to DPM with $\mathcal{S}^{\text{a}}$. (c) DPM-DQ can further boost the model performance. \textbf{Bolds} are the best results of DPM with $\mathcal{S}^{\text{a}}$ in corresponding setting. \underline{Underlines} are the best results of DPM with $\mathcal{S}^{\text{s}}$ in corresponding setting.}
\vspace{3pt}
\label{table:quantitative results in ablation study}
% \resizebox{0.48\textwidth}{!}{%
\begin{tabular}{@{}ccccccc@{}}
\Xhline{1pt}
 & $\mathcal{C}$ & $\mathcal{S}^{\text{a}}$  & $\mathcal{S}^{\text{s}}$ &   DPM-LQ & DPM-DQ & ACER(\%) $\downarrow$ \\ \hline
\multirow{3}{*}{(a)}  & \checkmark &    &    & & & 1.5\\
  & \checkmark & \checkmark &  &    &  & \textbf{1.1}\\  
 & \checkmark &  &  \checkmark &       &   & \underline{1.3} \\ \hline
\multirow{4}{*}{(b)}  & \checkmark & \checkmark &  & &  & 1.1\\   
  & \checkmark & \checkmark  &   &   \checkmark &  & \textbf{0.9} \\ 
   & \checkmark &  &  \checkmark &   &  & 1.3\\ 
 & \checkmark &  &  \checkmark  &   \checkmark &   & \underline{1.0} \\  \hline
\multirow{4}{*}{(c)}    & \checkmark & \checkmark  &   &   \checkmark &  & 0.9 \\  
 & \checkmark & \checkmark  &   &   \checkmark &  \checkmark  & \textbf{0.6} \\%\cmidrule{2-10}
  & \checkmark &  &  \checkmark  &   \checkmark & & 1.0 \\
  & \checkmark &  &  \checkmark  &   \checkmark & \checkmark & \underline{0.7} \\ \Xhline{1pt}
  %\bottomrule [\heavyrulewidth]
\end{tabular}%
% }
% \vspace{-12pt}
\end{table}

% \noindent \textbf{Feature Normalization is Indispensable.}
% Before DPM-DQ, we finetune the model through feature normalization. 
% %
% The reason is that Function Distance: $\omega\mu$ which is used in softmax is not proportional to the Euclidean Distance: $-(\omega - \mu)^{2}$ which used in Gaussian distribution. 
% %
% Therefore, without finetune, we can not sure the  larger $({\sigma^{\text{D}}})^{2}$ is related to the data with larger $D$, \textit{i.e.} False Reject and False Accept.
% %
% However, $\left \| \omega \right \|\left \| \mu \right \|$  is proportional to $-(\omega - \mu)^{2} $. 
% %
% As shown in Table~\ref{table:the results of finetune CelebA-Spoof}, the performance of DPM without finetuning becomes worse. It indicates that finetune is extremely important to data quality modeling learning. 

\noindent \textbf{Self-distributed Labels are Useful.}
Semantic information demonstrates successful auxiliary role in ~\cite{CelebA-Spoof}. 
We compare the role of self-distributed semantic label: $\mathcal{S}^{\text{s}}$ and annotated semantic label: $\mathcal{S}^{\text{a}}$. As shown in Table.~\ref{table:quantitative results in ablation study} (a), both $\mathcal{S}^{\text{s}}$ and $\mathcal{S}^{\text{a}}$ can improve the performance of baseline. However, it should be noted that there are only 28.5\% $\mathcal{S}^{\text{s}}$ that are equal to the corresponding $\mathcal{S}^{\text{a}}$. It indicates that even the distribution of $\mathcal{S}^{\text{s}}$ are not strictly the same as $\mathcal{S}^{\text{a}}$. $\mathcal{S}^{\text{s}}$ still distributes meaningful and implies useful semantic information.

\noindent \textbf{The Effectiveness of DPM-LQ.}
In Table.~\ref{table:quantitative results in ablation study} (b), for models with coarser labels like $\mathcal{S}^{\text{s}}$, DPM-LQ can play a greater role, and even stimulate $\mathcal{S}^{\text{s}}$ to play almost the same boosting role compared to $\mathcal{S}^{\text{a}}$. Specifically, after DPM-LQ, the model with coarser  $\mathcal{S}^{\text{s}}$ achieves comparable results compared to the model with $\mathcal{S}^{\text{a}}$. In other words, DPM-LQ stimulates the potential of $\mathcal{S}^{\text{s}}$ which is coarser. 

\noindent \textbf{Robustness to Noise.} 
In this part, based on $\mathcal{S}^{\text{s}}$,
we further deliberately change $\mathcal{S}^{\text{s}}$ of training data in $\mathcal{D}^{def}$: Oulu-NPU. 
Specifically, we randomly choose several certain percentages (20\%, 50\%, 70\%, and 100\%) of training data in $\mathcal{D}^{def}$, and randomly assign one type of self-distributed semantic spoof type labels to these samples. 
As shown in Fig.~\ref{figure:noise} (a), as label noise becomes extremely severe, DPM-LQ can barely work. 
The modeled data quality can stably improve performance no matter how noisy the semantic label is. 
Besides, we also conduct experiments to explore the influence of data noise in the test set. 
Specifically, we randomly select different proportions of samples from Oulu-NPU to pollute them with Gaussian blur. 
As shown in Fig.~\ref{figure:noise} (b), as the proportion of data noise becomes mid-level (20\%, 30\%), DPM-DQ can improve the performance of DPM more significantly. However, DPM-DQ also degrades when data noise becomes extremely serious.

\section{Conclusion}
\vspace{-3pt}
% In order to further improve the FAS performance by addressing the noise problem efficiently, 
In this work, we comprehensively study the noise problem in face anti-spoofing (FAS) for the first time.
We propose a clean yet powerful framework called Dual Probabilistic Modeling (DPM).
%which consists of two parts: 
%DPM-LQ and DPM-DQ.
The proposed DPM consists of two parts: 
DPM-LQ and DPM-DQ, both of which  are based on the assumption of coherent probabilistic distributions. 
Furthermore, we design the generalized DPM to tackle the problem of noisy labels and degraded data for practical use.
Extensive experimental results on standard benchmarks demonstrate the superiority of the proposed DPM over state-of-the-art methods.

% \setlength{\tabcolsep}{5pt}
% \begin{table}[t]
% \centering
% \ra{1.2}
% \caption{Notations}
% \label{table:notations}
% \resizebox{0.5\textwidth}{!}{%
% \begin{tabular}{@{}ll@{}}
% \toprule
% Notation & Meaning \\ \midrule
% $\mathcal{D}^{def}$ & Semantic Label Deficient Dataset \\
% $\mathcal{D}^{suf}$ & Semantic Label Sufficient Dataset \\
% $\mathcal{S}_{\texttt{s}}$ & Self-distributed Semantic Label \\
% $\mathcal{S}^{\text{s}}$ & Semantic Spoof Type Label \\
% $\mathcal{S}_{\texttt{s}}^{\text{s}}$ & Self-distributed Semantic Spoof Type Label \\
% $\mathcal{S}_{\texttt{a}}$ & Annotated Semantic Label \\
% $\mathcal{S}_{\texttt{a}}^{\text{s}}$ & Annotated Spoof Type Label \\ 
% $\mathcal{S}^{\text{i}}$ &Semantic Illumination Condition Label \\
% $\mathcal{S}_{\texttt{s}}^{\text{i}}$ & Self-distributed Semantic Illumination Condition Label \\ 
% $\mathcal{C}$ & Live/Spoof Label \\
% % $\mathcal{G}$ & Geometric Label \\
% \bottomrule
% \end{tabular}%
% }
% \end{table}

% \clearpage

\ifCLASSOPTIONcaptionsoff
  \newpage
\fi



%\bibliographystyle{IEEEtran}


{\small
\bibliographystyle{IEEEtran}
\bibliography{barejrnl}
}

\end{document}